%% file: main.tex
\documentclass[11pt]{article}
\usepackage[preprint]{acl}

\usepackage{times}
\usepackage{latexsym}
\usepackage[T1]{fontenc}
\usepackage[utf8]{inputenc}
\usepackage{microtype}

\usepackage{graphicx}

\usepackage{booktabs}
\usepackage{amsmath,amssymb,mathtools}
\usepackage{multirow}
\usepackage{makecell}
\usepackage{array}
\usepackage{tabularx}
\usepackage{xspace}
\usepackage{url}
\usepackage{algorithm}
\usepackage{float}
\usepackage{placeins}
\usepackage{algpseudocode}

\newcolumntype{Y}{>{\raggedright\arraybackslash}X}
\AtBeginDocument{%
  \setlength{\abovedisplayskip}{4pt plus 1pt minus 1pt}%
  \setlength{\belowdisplayskip}{4pt plus 1pt minus 1pt}%
  \setlength{\abovedisplayshortskip}{2pt plus 1pt minus 1pt}%
  \setlength{\belowdisplayshortskip}{3pt plus 1pt minus 1pt}%
  \setlength{\jot}{2pt}%
}

\newcommand{\method}{\textsc{CalibDCD}\xspace}
\newcommand{\fpp}{\mathrm{FPP}}
\newtheorem{example}{\textbf{Example}}
\title{Calibrating Post-Training Feature Shifts for LLM Data Contamination Detection}

\author{
Zhen Yang$^{1,*}$, Mengqi Wang$^{1,*}$, Gengda Zhao$^{1}$, Mo Zhou$^{1}$, Jianwei Wang$^{1}$, Wenjie Zhang$^{1}$ \\
$^{1}$The University of New South Wales
}

\makeatletter
\newcommand\blfootnote[1]{%
  \begingroup
  \renewcommand\thefootnote{}%
  \footnotetext{#1}%
  \addtocounter{footnote}{-1}%
  \endgroup
}
\makeatother

\begin{document}
\pagestyle{empty}
\maketitle
\thispagestyle{empty}
\blfootnote{\raggedright $^{*}$The first two authors contributed equally.\par Corresponding author: jianwei.wang1@unsw.edu.au}

\begin{abstract}
Large language models (LLMs) are trained on massive and largely undisclosed corpora that may contain copyrighted or privacy-sensitive content. Data contamination detection (DCD) therefore aims to determine whether a given text is a member of the pre-training corpus of a target LLM. Recent state-of-the-art DCD methods follow a feature-based paradigm that derives membership features from the input text and the corresponding model output. However, most modern LLMs undergo post-training, such as instruction tuning, preference optimization, and reasoning-oriented training, which can alter model outputs and shift the corresponding membership features, thereby reducing the separability between members and non-members.
To address this problem, we propose \method{}, a broadly applicable calibration framework for feature-based DCD methods, comprising (1) Multi-View Shift Detection, which identifies recurring feature shifts associated with post-training, and (2) Bounded Feature Correction, which selectively mitigates their influence on membership prediction.
Specifically, Multi-View Shift Detection evaluates controlled prompt variants on known non-member texts and consolidates the most informative views to identify recurring feature shifts. Bounded Feature Correction selectively adjusts feature components aligned with the detected shifts and controls the correction extent to preserve useful detection information.
Experiments show that \method{} consistently improves existing feature-based detectors, with gains of up to 7.0\% in AUC and 15.0\% in TPR@5\%FPR.
\end{abstract}

\input{sections/01_introduction}

\input{sections/02_related_work}

\input{sections/problem_statement}

\input{sections/03_method}

\input{sections/04_experiments}

\input{sections/05_analysis}
\FloatBarrier
\input{sections/06_conclusion}

\clearpage
\section*{Limitations}
\input{sections/07_limitations}

\section*{Ethics Statement}
\input{sections/09_ethics}

\bibliography{custom}

\appendix
\input{sections/08_appendix}

\end{document}

%% file: sections/01_introduction.tex
\section{Introduction}
\label{sec:introduction}

The pre-training corpora of large language models (LLMs) may contain copyrighted works, privacy-sensitive documents, benchmark examples, and other proprietary or sensitive materials \citep{brown2020language,carlini2021extracting,carlini2023quantifying,karamolegkou2023copyright,wang2025hse}.
Moreover, LLMs can memorize and reproduce portions of their pre-training data \citep{carlini2021extracting,carlini2023quantifying,jiang2026nsmem}, raising concerns about copyright, privacy, and evaluation integrity.
Data contamination detection (DCD) therefore aims to determine whether a given text was included in the pre-training corpus of a target LLM \citep{shi2024detecting,zhang2024pretraining,hu2025veilprobe}.
Following previous works, we refer to a text as a \emph{member} if it was included in this corpus and as a \emph{non-member} otherwise \citep{shokri2017membership,carlini2022membership,shi2024detecting}.

Existing DCD methods commonly operate in a black-box setting, in which the detector can query the target LLM and observe its generated outputs but has no access to model parameters or internal signals~\citep{ye2024data,dong2024generalization,deng2024investigating,golchin2025data}.
Early task-based methods formulate an input text as a task, such as cloze completion or multiple-choice selection, and infer membership from the response generated by the target model \citep{chang2023speak,duarte2024decop}.
Consequently, their effectiveness may vary with the task formulation and the ability of the target model to perform the constructed task.
Recent state-of-the-art (SOTA) feature-based methods, such as DPDLLM \citep{zhou2024dpdllm} and VeilProbe \citep{hu2025veilprobe}, derive \emph{membership features} from the input text and the corresponding output generated by the target LLM, and train a classifier to distinguish members from non-members.

However, the reliance of feature-based detectors makes them sensitive to behavioral changes introduced by post-training.
Modern LLMs commonly undergo post-training procedures, such as instruction tuning, preference optimization, and reasoning-oriented training, which can change the style, length, structure, and content of the generated output, consequently shifting the resulting membership features and reducing the separability between members and non-members~\citep{wei2022finetuned,ouyang2022training,bai2022constitutional,rafailov2023direct,deepseek2025r1}.

\begin{figure}[!t]
\centering
\includegraphics[width=\linewidth]{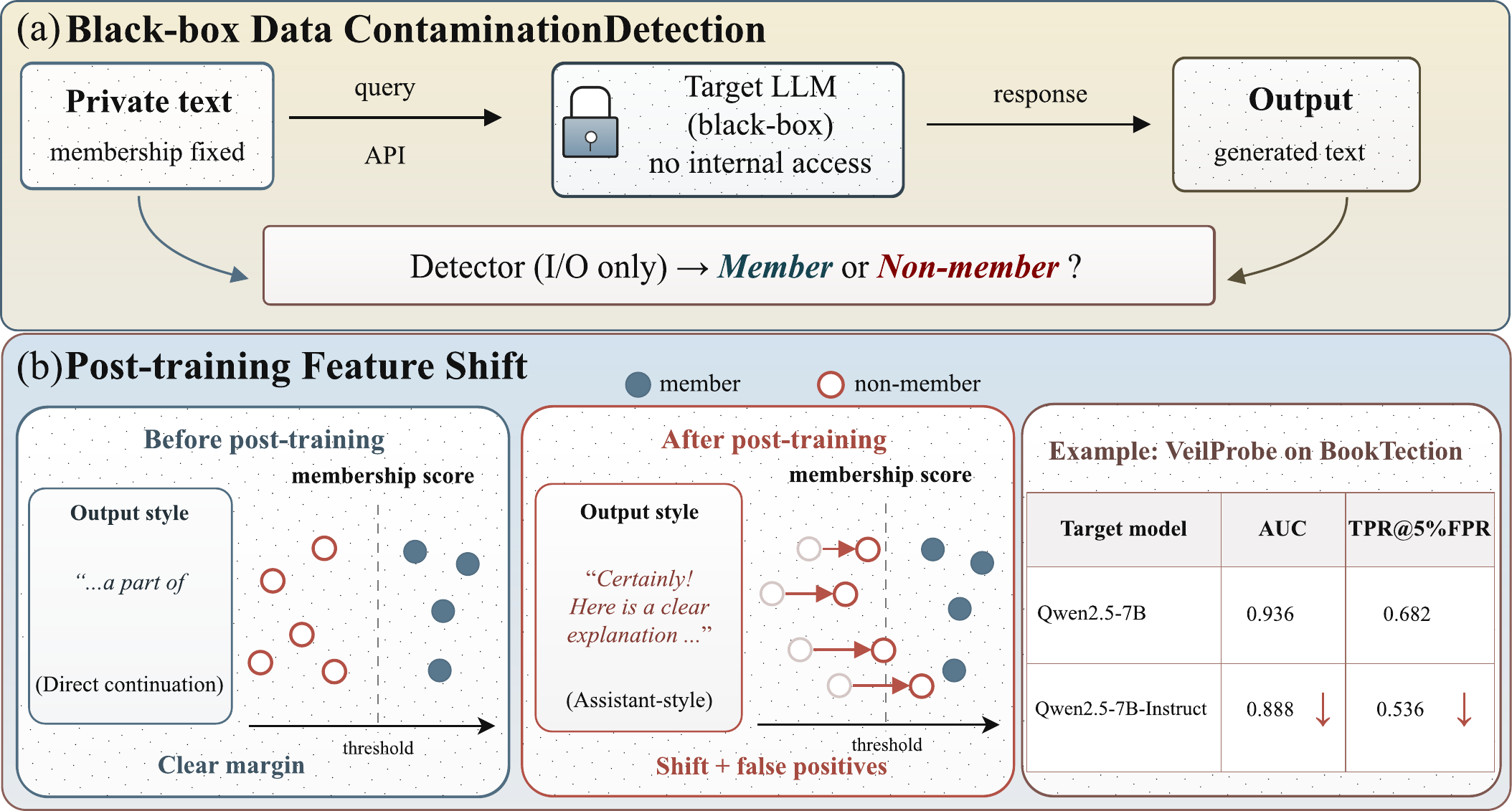}
\vspace{2pt}
\caption{Motivation of \method{}. (a) Black-box DCD. (b) Post-training-induced shifts and their impact on detection. The table compares VeilProbe on \textsc{BookTection} using the base and post-trained target models \citep{duarte2024decop,hu2025veilprobe}.}
\label{fig:motivation}
\end{figure}

\begin{example}
Figure~\ref{fig:motivation} illustrates how such shifts affect detection: assistant-style responses resulting from post-training increase the membership scores of non-members, causing some to cross the decision threshold and reducing the separation between members and non-members.
This effect is also evident in the VeilProbe results on \textsc{BookTection}: under otherwise identical settings, changing the target model from Qwen2.5-7B to its post-trained counterpart, Qwen2.5-7B-Instruct, decreases the DCD performance, with AUC dropping from 0.936 to 0.888 and TPR@5\%FPR from 0.682 to 0.536.
Appendix~\ref{app:base_vs_instruct} provides the complete comparison setup.
\end{example}

Two challenges remain in calibrating feature-based DCD against post-training-induced shifts:
(1) \textit{complex feature shifts under heterogeneous post-training}, as instruction tuning, preference optimization, and reasoning-oriented training can produce more instruction-following, preference-aligned, or step-by-step responses, altering output length, structure, lexical overlap, and continuation patterns in different directions and magnitudes~\citep{kirk2024understanding,zhou2023lima}; and
(2) \textit{balancing post-training shift correction and useful information preservation}, as shift-related feature components may also encode valid membership signals, making indiscriminate correction potentially detrimental~\citep{haghighatkhah2022better,belrose2023leace}.

To address these challenges, we propose \method{}, a broadly applicable calibration framework for feature-based DCD methods. It comprises (1) Multi-View Shift Detection, which evaluates multiple prompt variants on known non-member texts, prioritizes views based on their false-positive pressure (FPP), and establishes cross-view consensus to identify recurring feature-shift directions; and (2) Bounded Feature Correction, which selectively suppresses features aligned with the detected shift subspace while controlling the correction extent to limit the loss of useful detection information. The source code and experimental configurations are available at https://anonymous.4open.science/r/CALIBDCD/.

Our main contributions are:
\begin{itemize}
    \item We propose a broadly applicable calibration framework for mitigating the induced shifts.

    \item We develop Multi-View Shift Detection to rank prompt views by FPP and identify recurring shifts through cross-view consensus.

    \item We devise Bounded Feature Correction to selectively suppress shift-aligned features and preserve useful detection information.

    \item Extensive experiments show gains of up to 7.0\% in AUC and 15.0\% in TPR@5\%FPR.

\end{itemize}

%% file: sections/02_related_work.tex
\begin{figure*}[!t]
\centering
\includegraphics[width=0.98\textwidth]{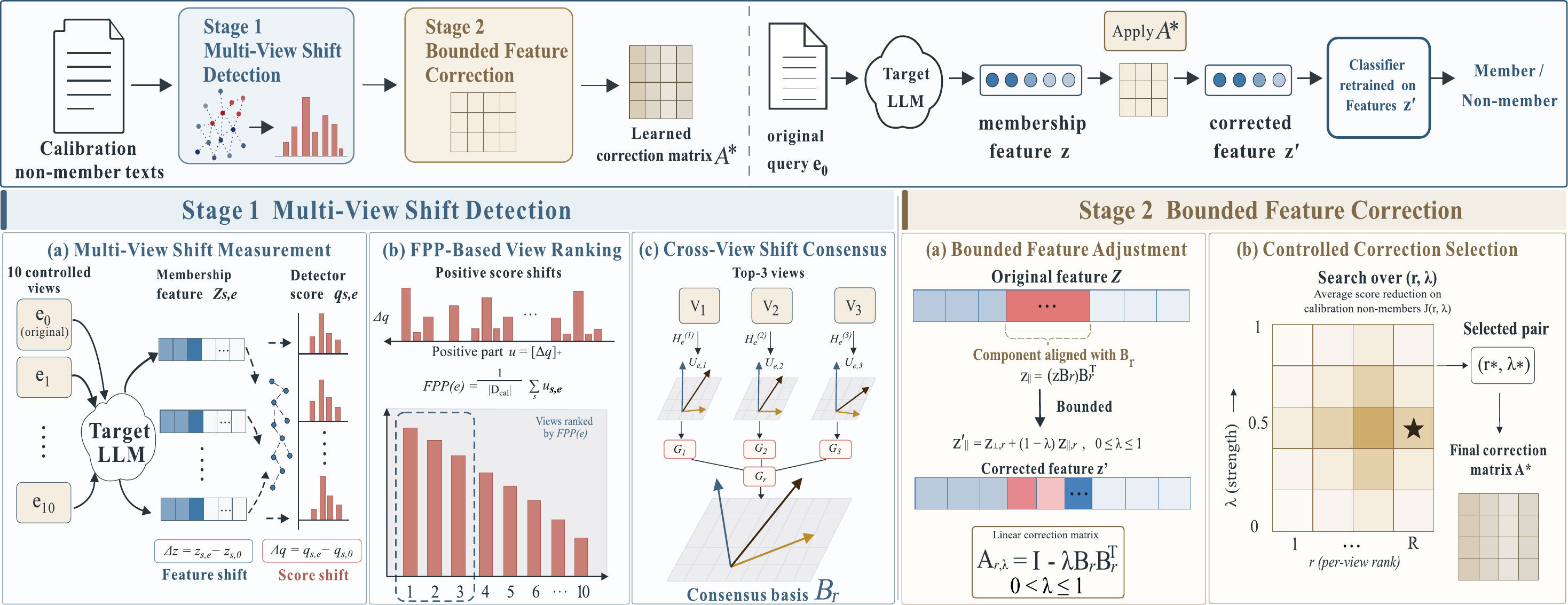}
\caption{Overview of \method{}, which combines Multi-View Shift Detection to estimate post-training-related feature-shift subspaces with Bounded Feature Correction to selectively attenuate shift-aligned components.}
\label{fig:overview}

\end{figure*}

\section{Related Work}
\label{sec:related}

\paragraph{Task-based Data Contamination Detection.}
One line of work formulates contamination detection as a behavioral task designed to reveal model familiarity with the input text.
Name-Cloze removes named entities from book passages and evaluates whether the target model can reconstruct the missing content~\citep{chang2023speak}, while DE-COP asks the model to distinguish an original passage from perturbed alternatives~\citep{duarte2024decop}.
Related studies investigate benchmark-level contamination by comparing model outputs across carefully constructed test variants~\citep{oren2023proving,maini2024llm}.
These methods are sensitive to task design and the task-solving ability of the target model.

\paragraph{Feature-based Data Contamination Detection.}
Feature-based methods convert observed model behavior into membership features and train a classifier to distinguish members from non-members.
Prior methods use likelihood, neighborhood, or divergence signals to measure how strongly a model fits an input text \citep{mattern2023membership,shi2024detecting,zhang2024pretraining}.
In the black-box setting, DPDLLM derives probability-based features from generated text using a reference language model \citep{zhou2024dpdllm}, while VeilProbe learns input--output mapping features and applies key-token perturbations \citep{hu2025veilprobe}.
Our work calibrates these output-dependent feature representations without changing their detector design.

%% file: sections/problem_statement.tex
\section{Problem Statement}
\label{subsec:detector_interface}

We first formally define the DCD task.

\noindent\textbf{Definition 1 (Data Contamination Detection).}
Let $\Theta$ denote a target LLM pre-trained on a corpus
$\mathcal D_{\Theta}$.
Given an input text $s$ and black-box access to $\Theta$, DCD aims to predict
\begin{equation}
m_s=\mathbb I\!\left[s\in\mathcal D_{\Theta}\right],
\end{equation}
where $m_s=1$ indicates that $s$ is a member of the pre-training corpus and $m_s=0$ otherwise.

A feature-based detector constructs
\begin{equation}
\mathbf z_{s,e_0}
=
\phi\bigl(e_0(s),\Theta(e_0(s))\bigr)
\in\mathbb R^d,
\label{eq:original_feature}
\end{equation}
where $e_0$ is the original query construction function, $\phi$ is the feature extractor, and $d$ is the feature dimension.
A scoring function $q$ maps $\mathbf z_{s,e_0}$ to a membership score, with a larger value indicating stronger membership evidence.
Given a threshold $\eta$, the predicted label is
\begin{equation}
\widehat m_s
=
\mathbb I\!\left[
q(\mathbf z_{s,e_0})\geq\eta
\right].
\end{equation}
To improve the robustness of feature-based DCD under post-training, we formulate the following feature calibration problem.

\noindent\textbf{Definition 2 (Feature Calibration under Post-Training).}
Let
$\mathcal D^-_{\mathrm{cal}}=\{s_i^-\}_{i=1}^{n}$
be a set of known non-members, where
$s_i^-\notin\mathcal D_{\Theta}$.
Given an existing feature-based detector, black-box access to $\Theta$,
and $\mathcal D^-_{\mathrm{cal}}$, feature calibration aims to determine
a transformation
\begin{equation}
\mathcal T(\cdot;\mathcal D^-_{\mathrm{cal}}):
\mathbb R^d\rightarrow\mathbb R^d
\end{equation}
that maps the original feature $\mathbf z_{s,e_0}$ to
\begin{equation}
\mathbf z'_{s,e_0}
=
\mathcal T\bigl(
\mathbf z_{s,e_0};
\mathcal D^-_{\mathrm{cal}}
\bigr),
\label{eq:calibrated_feature}
\end{equation}
with the objective of producing features that support more reliable
membership prediction under post-training-related shifts.
Appendix~\ref{app:notation} summarizes the main notation used in the formulation.

%% file: sections/03_method.tex
\section{Method}
\label{sec:method}

Figure~\ref{fig:overview} presents an overview of \method{}, a broadly applicable calibration framework that operates on the feature representations of existing feature-based DCD detectors, comprising (1) Multi-View Shift Detection, which estimates a consensus subspace of recurring post-training-related feature shifts using controlled prompt views, and (2) Bounded Feature Correction, which selectively attenuates components aligned with the consensus shift subspace by controlling the correction scope and strength.
Appendix~\ref{app:pipeline_walkthrough} provides an end-to-end pipeline walkthrough example with the complete calibration algorithm (Algorithm~\ref{alg:calibdcd}).

\subsection{Multi-View Shift Detection}
\label{subsec:shift_detection}

\noindent\textbf{Motivation.}
Multi-View Shift Detection aims to identify a stable subspace of post-training-related feature changes that are relevant to membership prediction. To achieve this goal, the module proceeds in three stages: (1) \textit{Multi-View Shift Measurement} evaluates known non-members under multiple controlled query variants and measures their feature changes relative to the original query to obtain complementary observations of post-training-related shifts; (2) \textit{FPP-Based View Ranking} prioritizes views that increase the membership scores of known non-members, thereby focusing subsequent analysis on shifts that are relevant to false-positive predictions; and (3) \textit{Cross-View Shift Consensus} retains directions consistently supported across the selected views, so that query-specific variations are excluded and recurring shift patterns form a stable subspace.

\paragraph{Multi-View Shift Measurement.}
\label{subsubsec:shift_measurement}
The original query provides only a single view of model behavior, limiting the characterization of post-training-related feature shifts across query conditions. \method{} therefore evaluates each known non-member under multiple controlled views and compares the resulting features and scores with those from the original query, providing complementary evidence for identifying stable, detector-relevant shift directions.

Specifically, \method{} maintains a bank $\mathcal E_{\mathrm{cand}}$ of controlled views, comprising eight universal response-format views and two assistant-generation-boundary views tailored to each target-model family.
Each view $e\in\mathcal E_{\mathrm{cand}}$ is a query construction function that maps an input text $s$ to a query variant $e(s)$ by modifying only the query prefix or generation boundary while preserving the text and its membership status.
The universal views add response cues such as \textit{Assistant}, \textit{Answer}, \textit{Reasoning}, and \textit{Summary} before the input text to elicit different response formats.
The two model-specific views vary the chat-template boundary between the user input and assistant generation.
For Qwen, the two views differ in whether they include an additional separator. For Llama, they use compact and official header formats, respectively. For DeepSeek, they differ in whether they include the reasoning opener (detailed in Appendix~\ref{app:view_bank}).

For each calibration non-member $s$ and controlled view $e$, the target model generates a response to $e(s)$, and the original detector computes
\begin{equation}
\mathbf z_{s,e}
=
\phi\bigl(e(s),\Theta(e(s))\bigr),
\qquad
q_{s,e}=q(\mathbf z_{s,e}).
\end{equation}
Using the feature $\mathbf z_{s,e_0}$ and score $q_{s,e_0}$ obtained under the original query $e_0$ as references, we define
\begin{align}
\Delta\mathbf z_{s,e}
&=\mathbf z_{s,e}-\mathbf z_{s,e_0},
\label{eq:feature_shift}\\
\Delta q_{s,e}
&=q_{s,e}-q_{s,e_0}.
\label{eq:score_shift}
\end{align}
Because the two observations share the same input text, target model, feature extractor, and scoring function, these paired differences isolate the changes exposed by the query view. $\Delta\mathbf z_{s,e}$ characterizes the feature displacement, while $\Delta q_{s,e}$ measures its effect on the membership score. Together, these paired changes provide the sample-level evidence used to rank controlled views and estimate score-relevant shift directions.

\paragraph{FPP-Based View Ranking.}
\label{subsubsec:fpp_ranking}
Not all controlled views expose feature changes that are equally relevant to membership prediction. Some views may induce substantial feature displacement without affecting detector scores, whereas views that consistently increase the scores of known non-members provide stronger evidence of false-positive risk. We therefore rank the candidate views using \emph{false-positive pressure} (FPP), which measures their average positive score increase over the calibration non-members.
We retain the positive part of each score change
\begin{equation}
u_{s,e}=[\Delta q_{s,e}]_+
=\max(0,\Delta q_{s,e}),
\label{eq:positive_score_weight}
\end{equation}
and define the FPP of view $e$ as
\begin{equation}
\fpp(e)=\frac{1}{|\mathcal D^-_{\mathrm{cal}}|}
\sum_{s\in\mathcal D^-_{\mathrm{cal}}}u_{s,e}, 
\label{eq:fpp}
\end{equation}
where a higher FPP indicates that a view induces larger aggregate positive score
changes among known non-members, making it more informative for identifying
feature changes associated with false-positive tendency.
\method{} retains the three highest-FPP views $\mathcal E_{\mathrm{sel}}$ from $\mathcal E_{\mathrm{cand}}$.

\paragraph{Cross-View Shift Consensus.}
\label{subsubsec:shift_consensus}

After FPP-based ranking identifies the views that most strongly shift known non-members toward member predictions, it remains necessary to determine which feature directions are associated with these score increases. Therefore, \method estimates a score-guided shift subspace for each selected view and aggregates these view-specific subspaces to retain directions with consistent cross-view support.

For each selected view $e$, the sample-level score increase $u_{s,e}$ is used to weight the corresponding feature shift $\Delta\mathbf z_{s,e}$. To limit the influence of extreme score changes, we define the within-view clipping cap as the $95$th percentile of the positive score increases:
\begin{equation}
c_e
=
\operatorname{Percentile}_{95}
\left(
\left\{
u_{s,e}
\mid
s\in\mathcal D^-_{\mathrm{cal}},
\ u_{s,e}>0
\right\}
\right).
\label{eq:clipping_cap}
\end{equation}
The corresponding clipped weight and weighted feature shift are
\begin{equation}
w_{s,e}=\min(u_{s,e},c_e),
\qquad
\mathbf h_{s,e}=\sqrt{w_{s,e}}\,\Delta\mathbf z_{s,e}.
\label{eq:weighted_shift}
\end{equation}
Stacking $\mathbf h_{s,e}$ as the rows of $\mathbf H_e$ gives
\begin{equation}
\mathbf H_e^{\top}\mathbf H_e
=
\sum_{s\in\mathcal D^-_{\mathrm{cal}}}
w_{s,e}\,
\Delta\mathbf z_{s,e}^{\top}\Delta\mathbf z_{s,e}.
\label{eq:weighted_scatter}
\end{equation}
Thus, each sample contributes in proportion to its clipped positive score increase. The weighted shifts are not centered because their common directed component is part of the recurring score-increasing movement being estimated. We then compute
\begin{equation}
\mathbf H_e
=
\mathbf P_e\mathbf\Sigma_e\mathbf U_e^{\top},
\label{eq:view_svd}
\end{equation}
and let $\mathbf U_{e,r}$ contain the first $r$ right singular vectors. Its columns span the dominant score-guided feature-shift subspace for view $e$, where $r$ controls the complexity of the view-specific estimate.

Because a direction estimated from a single view may reflect query-specific variation rather than a recurring shift, the view-specific subspaces are combined through the average projector
\begin{equation}
\mathbf G_r
=
\frac{1}{|\mathcal E_{\mathrm{sel}}|}
\sum_{e\in\mathcal E_{\mathrm{sel}}}
\mathbf U_{e,r}\mathbf U_{e,r}^{\top}.
\label{eq:consensus_operator}
\end{equation}
Projector averaging gives each selected view equal influence after its score-guided subspace has been estimated, avoiding domination by a single view in a pooled decomposition. We then compute
\begin{align}
\mathbf G_r
&=
\mathbf V_r\,
\mathrm{diag}(\gamma_{r,1},\ldots,\gamma_{r,d})
\mathbf V_r^{\top},
\label{eq:consensus_eigendecomposition}\\
\gamma_{r,i}
&=
\frac{1}{|\mathcal E_{\mathrm{sel}}|}
\sum_{e\in\mathcal E_{\mathrm{sel}}}
\left\|
\mathbf U_{e,r}^{\top}\mathbf v_{r,i}
\right\|_2^2,
\label{eq:consensus_score}
\end{align}
where $\mathbf v_{r,i}$ is the $i$-th eigenvector of $\mathbf G_r$. Since $\gamma_{r,i}\in[0,1]$ measures the average squared projection of $\mathbf v_{r,i}$ onto the selected view subspaces, it quantifies the cross-view support for that direction. Given a support threshold $\tau$, the consensus basis is defined as
\begin{equation}
\mathbf B_r
=
\bigl[
\mathbf v_{r,i}
\mid
\gamma_{r,i}\geq\tau
\bigr].
\label{eq:consensus_basis}
\end{equation}
Here, $r$ controls the complexity of each view-specific subspace, whereas $\tau$ controls the degree of cross-view consistency required for a direction to be retained. Appendix~\ref{app:method_properties} provides the range argument for $\gamma_{r,i}$.

\subsection{Bounded Feature Correction}
\label{subsec:correction}
\noindent\textbf{Motivation.}
Bounded Feature Correction aims to reduce the influence of the detected shift subspace while limiting unnecessary modification of potentially useful feature components. Directly removing all shift-aligned components is undesirable because the detected directions may contain both post-training-related variation and valid membership evidence. To balance these effects, the stage proceeds in two steps: (1) \textit{Bounded Feature Adjustment} constructs candidate corrections that attenuate, rather than completely remove, feature components aligned with the consensus shift directions; and (2) \textit{Controlled Correction Selection} determines the correction scope and strength using known non-members, so that detector scores are reduced without applying an unnecessarily aggressive transformation.

\paragraph{Bounded Feature Adjustment.}
\label{subsubsec:bounded_adjustment}
The consensus basis identifies recurring score-increasing directions, but these directions may also contain information useful for membership prediction. To avoid indiscriminate removal, \method attenuates only the component aligned with the consensus subspace while preserving its orthogonal complement. Formally, because the columns of $\mathbf B_r$ are orthonormal, an original-format feature $\mathbf z$ can be decomposed as
\begin{align}
\mathbf z_{\parallel,r}
&=(\mathbf z\mathbf B_r)\mathbf B_r^{\top},
\label{eq:aligned_component}\\
\mathbf z_{\perp,r}
&=\mathbf z-\mathbf z_{\parallel,r}.
\label{eq:orthogonal_component}
\end{align}
For rank $r$ and strength $\lambda$, we construct
\begin{equation}
\begin{aligned}
\mathbf A_{r,\lambda}
&=\mathbf I-\lambda\mathbf B_r\mathbf B_r^{\top},
\qquad 0\leq\lambda\leq1,\\
\mathbf z'
&=\mathbf z\mathbf A_{r,\lambda}
=\mathbf z_{\perp,r}+(1-\lambda)\mathbf z_{\parallel,r}.
\end{aligned}
\label{eq:corrected_feature}
\end{equation}
The operator acts only on the consensus subspace
\begin{equation}
\begin{aligned}
\mathbf A_{r,\lambda}\mathbf B_r
&=(1-\lambda)\mathbf B_r,\\
\mathbf A_{r,\lambda}\mathbf v
&=\mathbf v
\quad\text{if }\mathbf B_r^{\top}\mathbf v=\mathbf 0.
\end{aligned}
\label{eq:bounded_action}
\end{equation}
The transformation is restricted to the consensus subspace, leaving all orthogonal components unchanged. The parameter $\lambda$ controls the attenuation strength, ranging from no correction at $\lambda=0$ to complete removal of the aligned component at $\lambda=1$, thereby allowing partial correction when full removal may discard useful information.

\paragraph{Controlled Correction Selection.}
\label{subsubsec:controlled_selection}

Using the candidate correction matrices $\mathbf A_{r,\lambda}$ constructed above, \method selects the final correction based only on known non-members. Let $\mathcal R$ and $\Lambda$ denote the predefined candidate sets for the subspace rank and attenuation strength, respectively. The original detector is first trained to obtain the scoring function $q$, which remains fixed during correction selection. For each candidate pair $(r,\lambda)$, its calibration objective is defined as the average reduction in non-member scores:
\begin{equation}
J(r,\lambda)
=
\frac{1}{|\mathcal D^-_{\mathrm{cal}}|}
\sum_{s\in\mathcal D^-_{\mathrm{cal}}}
\left[
q(\mathbf z_{s,e_0})
-
q\bigl(\mathbf z_{s,e_0}\mathbf A_{r,\lambda}\bigr)
\right].
\label{eq:parameter_selection}
\end{equation}
A larger $J(r,\lambda)$ indicates that the candidate correction more strongly reduces the membership evidence assigned to known non-members. The final parameters and correction matrix are selected as
\begin{align}
(r^{*},\lambda^{*})
&=
\arg\max_{r\in\mathcal R,\,\lambda\in\Lambda}
J(r,\lambda),
\label{eq:selected_parameters}\\
\mathbf A^{*}
&=
\mathbf A_{r^{*},\lambda^{*}}.
\label{eq:selected_correction}
\end{align}

After selecting $\mathbf A^{*}$, the same correction is applied to the
supervised training features and to the features of each input text before
classification. A classifier from the original detector family is then
retrained on the corrected training features, yielding the final scoring
function $q_{\mathrm{final}}$. For an input text $s$, the calibrated detector
computes
\begin{equation}
\mathbf z'_{s,e_0}
=
\mathcal T_{\mathbf A^{*}}(\mathbf z_{s,e_0}),
\qquad
\widetilde q(s)
=
q_{\mathrm{final}}(\mathbf z'_{s,e_0}),
\label{eq:post_calibration_score}
\end{equation}
where $\mathcal T_{\mathbf A^{*}}$ denotes the detector-compatible application
of the selected correction.

%% file: sections/04_experiments.tex
\section{Experiments}
\label{sec:experiments}

\subsection{Experimental Setup}
\noindent \textbf{Datasets.}
Following prior work, we evaluate \method{} on four pre-training data
detection benchmarks. \textsc{WikiMIA} consists of Wikipedia event snippets,
while \textsc{BookMIA} contains book passages
\citep{shi2024detecting}. \textsc{BookTection} contains excerpts from
copyrighted books, and \textsc{ArxivTection} contains passages from
scientific papers on arXiv
\citep{duarte2024decop}. Table~\ref{tab:data_protocol} summarizes the
dataset statistics and split sizes.

\begin{table}[!t]
\centering
\scriptsize
\setlength{\tabcolsep}{4pt}
\renewcommand{\arraystretch}{1.05}
\caption{Statistics of the DCD benchmarks. Parentheses report the numbers of members and non-members.}
\label{tab:data_protocol}
\begin{tabular}{lrrr}
\toprule
\textbf{Benchmark}
& \textbf{Total}
& \makecell{\textbf{Supervised Train}}
& \makecell{\textbf{Evaluation Pool}} \\
\midrule
\textsc{BookTection}
& $2{,}000$
& $100\;(50/50)$
& $1{,}900\;(950/950)$ \\

\textsc{BookMIA}
& $2{,}000$
& $100\;(50/50)$
& $1{,}900\;(950/950)$ \\

\textsc{ArxivTection}
& $1{,}548$
& $100\;(50/50)$
& $1{,}448\;(712/736)$ \\

\textsc{WikiMIA}
& $542$
& $100\;(50/50)$
& $442\;(234/208)$ \\
\bottomrule
\end{tabular}
\end{table}

\begin{table*}[t]
\centering
\scriptsize
\setlength{\tabcolsep}{2.0pt}
\renewcommand{\arraystretch}{1.08}
\caption{AUC results of VeilProbe and DPDLLM with and without \method across four benchmarks and three target LLMs.
Bold values indicate the better result within each baseline--calibrated pair.}
\label{tab:auc_main}
\begin{tabular*}{\textwidth}{@{\extracolsep{\fill}}lcccccccccccc@{}}
\toprule
\multirow{2}{*}{\textbf{Method}} & \multicolumn{3}{c}{\textbf{BookTection}} & \multicolumn{3}{c}{\textbf{BookMIA}} & \multicolumn{3}{c}{\textbf{ArxivTection}} & \multicolumn{3}{c}{\textbf{WikiMIA}} \\
\cmidrule(lr){2-4}\cmidrule(lr){5-7}\cmidrule(lr){8-10}\cmidrule(lr){11-13}
& Qwen & Llama & \makecell{Deep-\\Seek} & Qwen & Llama & \makecell{Deep-\\Seek} & Qwen & Llama & \makecell{Deep-\\Seek} & Qwen & Llama & \makecell{Deep-\\Seek} \\
\midrule
VeilProbe & 88.8 & 88.2 & 86.8 & 78.5 & 80.7 & 77.0 & 91.3 & 90.3 & 86.0 & 94.6 & 95.1 & 93.9 \\
VeilProbe $+$ \method & \textbf{91.7} & \textbf{90.6} & \textbf{89.1} & \textbf{84.0} & \textbf{84.4} & \textbf{79.3} & \textbf{93.4} & \textbf{91.4} & \textbf{87.5} & \textbf{96.1} & \textbf{95.2} & \textbf{95.5} \\
\midrule
DPDLLM & 72.5 & 73.5 & 70.3 & 67.3 & 75.8 & 54.3 & 64.5 & 58.7 & 53.5 & 61.2 & 66.6 & 57.0 \\
DPDLLM $+$ \method & \textbf{74.2} & \textbf{75.8} & \textbf{71.4} & \textbf{70.0} & \textbf{78.5} & \textbf{61.3} & \textbf{65.8} & \textbf{59.3} & \textbf{55.1} & \textbf{62.4} & \textbf{67.4} & \textbf{57.8} \\
\bottomrule
\end{tabular*}
\end{table*}

\begin{table*}[t]
\centering
\scriptsize
\setlength{\tabcolsep}{2.0pt}
\renewcommand{\arraystretch}{1.08}
\caption{TPR@5\%FPR results of VeilProbe and DPDLLM with and without \method across four benchmarks and three target LLMs.
Bold values indicate the better result within each baseline--calibrated pair.}
\label{tab:tpr_main}
\begin{tabular*}{\textwidth}{@{\extracolsep{\fill}}lcccccccccccc@{}}
\toprule
\multirow{2}{*}{\textbf{Method}} & \multicolumn{3}{c}{\textbf{BookTection}} & \multicolumn{3}{c}{\textbf{BookMIA}} & \multicolumn{3}{c}{\textbf{ArxivTection}} & \multicolumn{3}{c}{\textbf{WikiMIA}} \\
\cmidrule(lr){2-4}\cmidrule(lr){5-7}\cmidrule(lr){8-10}\cmidrule(lr){11-13}
& Qwen & Llama & \makecell{Deep-\\Seek} & Qwen & Llama & \makecell{Deep-\\Seek} & Qwen & Llama & \makecell{Deep-\\Seek} & Qwen & Llama & \makecell{Deep-\\Seek} \\
\midrule
VeilProbe & 53.6 & 60.3 & 45.3 & 41.5 & 45.2 & 36.7 & 63.1 & 63.8 & 50.4 & 74.8 & 70.1 & 69.7 \\
VeilProbe $+$ \method & \textbf{68.6} & \textbf{66.1} & \textbf{58.0} & \textbf{49.8} & \textbf{55.7} & \textbf{41.6} & \textbf{72.2} & \textbf{68.4} & \textbf{57.3} & \textbf{81.6} & \textbf{71.4} & \textbf{75.6} \\
\midrule
DPDLLM & 15.3 & 22.7 & 12.4 & 17.9 & 20.2 & 6.6 & 15.9 & 8.3 & 5.9 & 12.4 & 13.7 & 8.5 \\
DPDLLM $+$ \method & \textbf{26.0} & \textbf{30.0} & \textbf{19.3} & \textbf{22.2} & \textbf{32.2} & \textbf{15.5} & \textbf{16.4} & \textbf{12.9} & \textbf{9.6} & \textbf{20.1} & \textbf{20.1} & \textbf{12.4} \\
\bottomrule
\end{tabular*}
\end{table*}

\noindent \textbf{Target LLMs.}
We evaluate Qwen2.5-7B-Instruct~\citep{qwen2_5},
Llama-3.1-8B-Instruct~\citep{dubey2024llama3}, and
DeepSeek-R1-Distill-Qwen-7B~\citep{deepseek2025r1}. These models are
comparable in scale but differ in model family and post-training design,
allowing us to evaluate across diverse post-training settings.

\noindent\textbf{Evaluation Metrics.}
Following prior works~\citep{hu2025veilprobe, carlini2022membership}, we use the Area Under the Receiver Operating
Characteristic Curve (AUC) and the True Positive Rate at a 5\% False Positive
Rate (TPR@5\%FPR) as evaluation metrics. A higher AUC indicates better overall
discrimination between members and non-members across classification
thresholds, while a higher TPR@5\%FPR indicates stronger member detection
under a low false-positive constraint.

\noindent \textbf{Baselines.}
We apply \method{} to two feature-based DCD detectors:
\textsc{VeilProbe}~\citep{hu2025veilprobe}, which learns input--output mapping
features and applies key-token perturbations, and
\textsc{DPDLLM}~\citep{zhou2024dpdllm}, which derives probability-based
features from generated text using a reference language model.

\noindent\textbf{Implementation Details.}
For each setting, we construct ten fixed controlled views, including eight
universal views and two model-specific views, from which FPP selects the top
three. We set $\mathcal R=\{3,4,5,6\}$, $\Lambda=\{0.7,0.8,0.9,1.0\}$, and
$\tau=0.95$. Calibration uses only known non-members and does not access evaluation metrics or member labels from the evaluation pool. Controlled-view outputs, features, and scores are cached and reused throughout calibration. For \textsc{VeilProbe}, correction is applied only to the output-derived feature block, whereas for \textsc{DPDLLM}, it is applied to the complete membership-feature vector. After correction, the original classifier family is retrained on the corrected supervised training features. Appendix~\ref{app:hyperparameters} provides the complete configuration. Experiments are conducted using Python~3.10 and CUDA~12.6 on a server equipped with eight NVIDIA RTX A5000 GPUs with $24$~GB of memory each.

\subsection{Main Results}
\label{subsec:main_results}

Tables~\ref{tab:auc_main} and~\ref{tab:tpr_main} show that \method{} improves both AUC and TPR@5\%FPR in all 24 settings, with average gains of 2.1\% and 7.0\%, respectively. The largest gains reach 7.0\% in AUC for DPDLLM on BookMIA with DeepSeek and 15.0\% in TPR@5\%FPR for VeilProbe on BookTection with Qwen, demonstrating consistent effectiveness across benchmarks, target models, and detector families.

The improvement is particularly pronounced under the low-FPR condition. VeilProbe gains 2.3\% in AUC and 7.7\% in TPR@5\%FPR on average, while DPDLLM gains 2.0\% and 6.4\%, respectively. This trend is consistent with \method{} focusing on score-increasing shifts of non-members, which mainly reduces high-confidence false positives. Across Qwen, Llama, and DeepSeek, the respective AUC gains are 2.4\%, 1.7\%, and 2.3\%, confirming robustness to different post-training designs. At the benchmark level, BookMIA achieves the largest average AUC gain of 4.0\%, whereas BookTection obtains the largest TPR@5\%FPR gain of 9.7\%, indicating that \method{} adapts to different types of post-training-induced feature shifts.

%% file: sections/05_analysis.tex
\begin{table}[!t]
\centering
\scriptsize
\setlength{\tabcolsep}{3.0pt}
\renewcommand{\arraystretch}{1.05}
\caption{Ablation results in percentages. Parentheses show the average change from the complete method on the same settings.}
\label{tab:ablation}
\begin{tabular}{lrrr}
\toprule
\textbf{Configuration} & \textbf{$n$} & \textbf{AUC} & \textbf{TPR@5\%FPR} \\
\midrule
\method & $24$ & $78.3\%$ & $41.8\%$ \\
Random-3 Views & $24$ & $77.7\%\;(-0.6\%)$ & $38.7\%\;(-3.1\%)$ \\
Top-1 View & $24$ & $77.2\%\;(-1.1\%)$ & $37.6\%\;(-4.2\%)$ \\
No Cross-View Consensus & $24$ & $77.0\%\;(-1.3\%)$ & $37.9\%\;(-3.9\%)$ \\
Binary Positive Weight & $24$ & $77.4\%\;(-0.9\%)$ & $38.8\%\;(-3.0\%)$ \\
\midrule
\method ($\lambda<1$) & $18$ & $78.0\%$ & $40.6\%$ \\
Fixed Full Correction & $18$ & $77.1\%\;(-0.9\%)$ & $38.0\%\;(-2.6\%)$ \\
\bottomrule
\end{tabular}
\end{table}

\begin{figure}[!t]
\centering
\includegraphics[width=\linewidth]{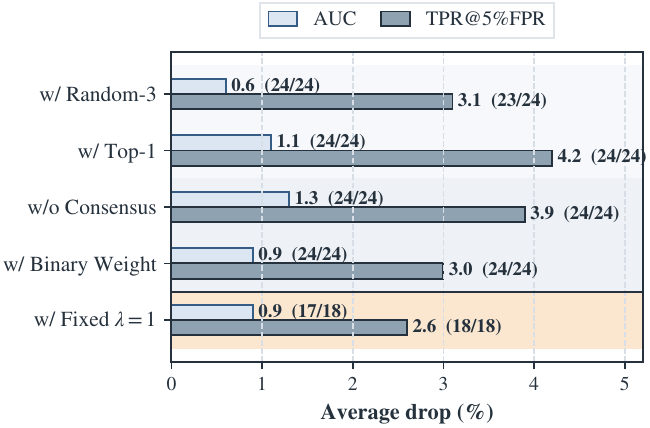}
\caption{Average performance drop relative to the complete method. Parentheses report the number of settings in which the complete method performs better. The final comparison includes the $18$ settings with a selected $\lambda<1$.}
\label{fig:ablation_summary}
\end{figure}

\subsection{Ablation Studies}
We change one component at a time while fixing all other settings. The variants replace FPP-based view selection with seeded Random-3 views, retain only the top-FPP view, remove cross-view consensus, use binary positive weights, or apply full correction with $\lambda=1$ to the $18$ settings where \method{} selects $\lambda<1$. Table~\ref{tab:ablation} and Figure~\ref{fig:ablation_summary} report the absolute results and average performance drops.

Replacing FPP-based selection with Random-3 Views reduces AUC and TPR@5\%FPR by $0.6\%$ and $3.1\%$, respectively, with \method{} achieving higher AUC in all $24$ settings and higher TPR in $23$. This confirms that prioritizing views with stronger false-positive pressure provides more informative shift observations than arbitrary view selection. Using only the Top-1 View causes larger drops of $1.1\%$ in AUC and $4.2\%$ in TPR@5\%FPR, and underperforms \method{} in all $24$ settings on both metrics, demonstrating the importance of combining complementary views rather than relying on a single view.

Removing Cross-View Consensus produces the largest AUC reduction of $1.3\%$ and decreases TPR@5\%FPR by $3.9\%$ across all $24$ settings. This shows that consensus filtering is necessary to retain recurring shift directions while excluding view-specific variations. Replacing score-sensitive weights with Binary Positive Weight reduces AUC by $0.9\%$ and TPR@5\%FPR by $3.0\%$ in every setting, indicating that weighting feature shifts by their induced score increases better captures their contribution to false-positive behavior. Finally, Fixed Full Correction reduces AUC by $0.9\%$ and TPR@5\%FPR by $2.6\%$ across the $18$ partial-correction settings; the selected strengths perform better in $17$ settings for AUC and all $18$ for TPR. This validates the importance of bounded correction in mitigating shift-related components without unnecessarily removing useful membership information.

\begin{figure}[!t]
\centering
\includegraphics[width=\linewidth]{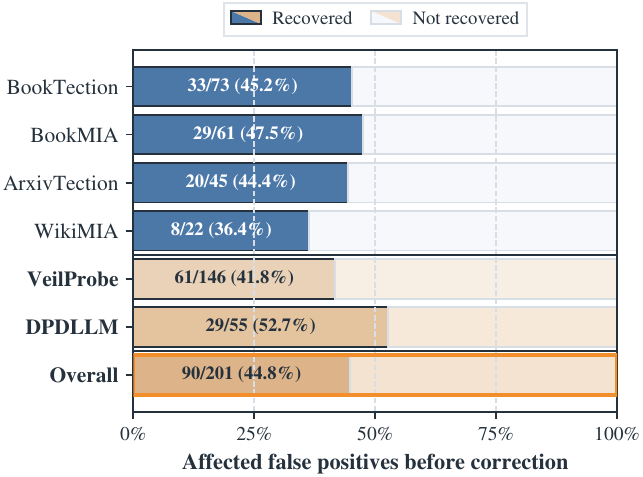}
\caption{Outcomes for affected false positives at the corrected detector's $5\%$ FPR threshold. Benchmark and detector rows partition the same $201$ cases. Labels report the number and proportion of recovered cases among affected cases.}
\label{fig:fp_recovery}
\end{figure}

\subsection{Analysis}
\noindent \textbf{Recovery of Shift-Associated False Positives.}
To examine whether \method{} improves the specific decisions targeted by
calibration, we conduct a decision-level recovery analysis. For each
benchmark--model--detector setting, we consider the selected controlled view
with the highest FPP and identify the known non-members associated with this
view that are misclassified as members by the original detector at its
$5\%$ FPR operating point. We then determine whether the corrected detector
reclassifies these examples as non-members using its corresponding threshold
at the same $5\%$ FPR, thereby holding the operating constraint constant.
Figure~\ref{fig:fp_recovery} reports the recovery rates by benchmark and
detector.
Overall, $90$ of the $201$ identified false positives are recovered as true
negatives, corresponding to a recovery rate of $44.8\%$. The recovery rate
ranges from $36.4\%$ on \textsc{WikiMIA} to $47.5\%$ on
\textsc{BookMIA}, and averages $41.8\%$ for \textsc{VeilProbe} and
$52.7\%$ for \textsc{DPDLLM}. These results show that \method{} mitigates a
substantial fraction of the false positives emphasized by high-FPP views,
providing decision-level evidence that complements the aggregate improvements
in AUC and TPR@5\%FPR.

\begin{figure}[!t]
    \centering
    \includegraphics[width=\columnwidth]{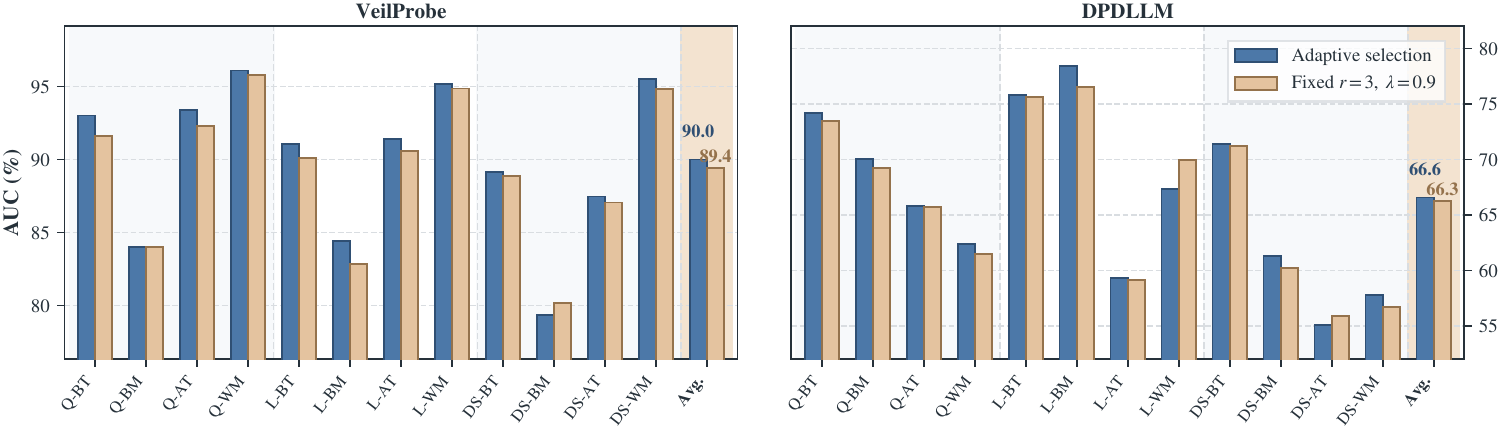}
    \caption{AUC comparison between adaptive correction selection and the
    fixed configuration $(r=3,\lambda=0.9)$. Q, L, and DS denote Qwen,
    Llama, and DeepSeek, while BT, BM, AT, and WM denote
    \textsc{BookTection}, \textsc{BookMIA}, \textsc{ArxivTection}, and
    \textsc{WikiMIA}, respectively.}
    \label{fig:adaptive_auc}
\end{figure}

\begin{figure}[!t]
    \centering
    \includegraphics[width=\columnwidth]{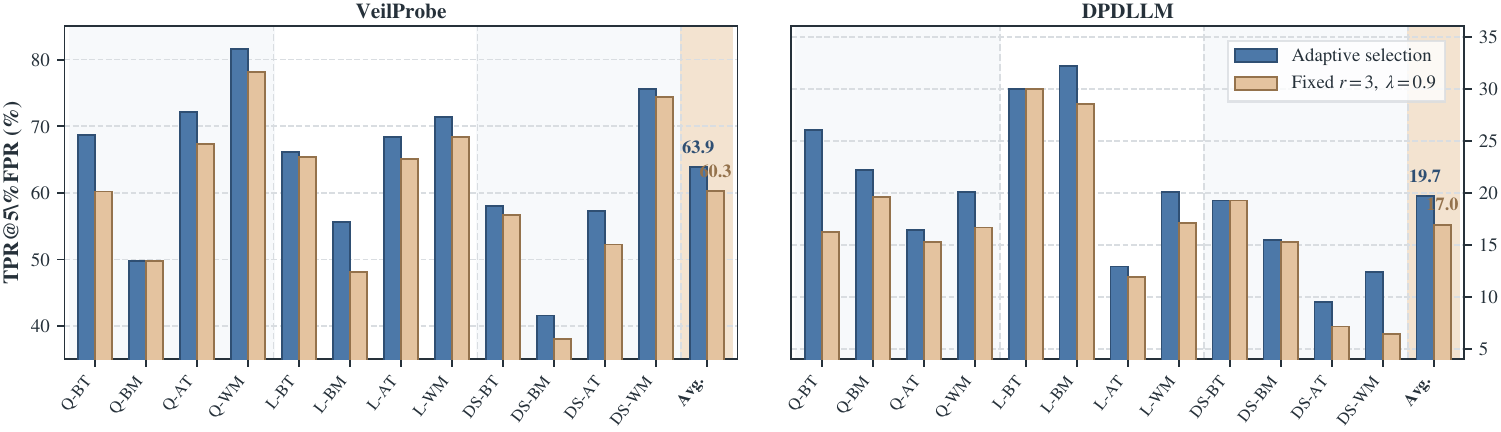}
    \caption{TPR@5\%FPR comparison between adaptive correction selection and
    the fixed configuration $(r=3,\lambda=0.9)$. Abbreviations follow
    Figure~\ref{fig:adaptive_auc}.}
    \label{fig:adaptive_tpr}
\end{figure}

\noindent\textbf{Adaptive Correction Selection.}
Figures~\ref{fig:adaptive_auc} and~\ref{fig:adaptive_tpr} compare adaptive
correction selection with the fixed configuration$(r=3,\lambda=0.9)$ in
terms of AUC and TPR@5\%FPR, respectively. Adaptive selection improves the average AUC from $89.4\%$ to $90.0\%$ for \textsc{VeilProbe} and from $66.3\%$ to $66.6\%$ for \textsc{DPDLLM}. The advantage is more pronounced at low FPR, where the average TPR@5\%FPR increases from $60.3\%$ to $63.9\%$ and from $17.0\%$ to $19.7\%$, respectively. These results indicate that adapting the correction rank and strength to each setting is more effective than using a single fixed configuration.

%% file: sections/06_conclusion.tex
\section{Conclusion}
\label{sec:conclusion}

We studied feature-based black-box data contamination detection under
post-training-induced output shifts. We proposed \method{}, which identifies
recurring score-increasing feature shifts from controlled views of known
non-members and applies a bounded correction before classification while
retaining the original detector query. Across four benchmarks, three
post-trained LLMs, and two detector families, \method{} improves both AUC and
TPR@5\%FPR in all $24$ settings, with average gains of $2.1\%$ and $7.0\%$
and maximum gains of $7.0\%$ and $15.0\%$,
respectively, without modifying the target LLM or detection-time query process.
These results highlight calibration as a practical way to improve the
robustness of feature-based DCD under post-training procedures.

%% file: sections/07_limitations.tex
\method corrects score-increasing feature shifts observed on known non-members.
Although post-training motivates the study, these shifts may also reflect dataset artifacts, decoding behavior, or the feature extractor.
The method therefore identifies shifts associated with higher non-member scores during calibration rather than proving that every corrected shift is caused by post-training.

\method requires known non-members outside supervised classifier training.
Temporal benchmarks and data released after a model cutoff provide such examples, but the requirement is harder to satisfy when the cutoff is unknown or no trusted non-member pool is available.
The calibration examples remain in the reporting pool in our protocol.
They provide non-member shift estimates, while reporting-pool member labels and evaluation metrics are not used to select the correction.
Calibration also requires additional target-model queries under controlled views, and its quality depends on whether these views expose the score-increasing shifts relevant to the detector.

The method applies to feature-based detectors that expose numerical membership features and scores.
Its bounded linear correction may not address nonlinear or sample-specific false-positive sources.

%% file: sections/09_ethics.tex
Data contamination detection can support audits of copyrighted, private, or benchmark data. Its outputs are probabilistic and should not be treated as definitive proof that a particular text was used in training. Audits should use lawfully obtained data and avoid unnecessary exposure of sensitive content.

%% file: sections/08_appendix.tex
\setcounter{topnumber}{4}
\setcounter{bottomnumber}{2}
\setcounter{totalnumber}{6}
\renewcommand{\topfraction}{0.95}
\renewcommand{\bottomfraction}{0.90}
\renewcommand{\textfraction}{0.05}
\renewcommand{\floatpagefraction}{0.80}
\setcounter{dbltopnumber}{2}
\renewcommand{\dbltopfraction}{0.95}
\renewcommand{\dblfloatpagefraction}{0.80}

\section{Notation}
\label{app:notation}

Table~\ref{tab:notation} summarizes the notation used in the main text.

\begin{table}[!htbp]
\centering
\small
\caption{Main notation.}
\label{tab:notation}
\begin{tabularx}{\linewidth}{lY}
\toprule
\textbf{Notation} & \textbf{Meaning} \\
\midrule
$\Theta$ & Target LLM. \\
$\mathcal{D}_{\Theta}$ & Pre-training corpus of $\Theta$. \\
$s$ & Input text to be tested. \\
$e_0$ & Original query used by the detector. \\
$e$ & Controlled view used during calibration. \\
$\phi$ & Membership feature extractor. \\
$\mathbf z_{s,e}\in\mathbb R^d$ & Membership feature under view $e$. \\
$q_{s,e}=q(\mathbf z_{s,e})$ & Detector score under view $e$. The scoring function $q$ remains fixed during calibration. \\
$q_{\mathrm{final}}$ & Final classifier score in the corrected feature space. \\
$\mathcal T_{\mathbf A^*}$ & Detector-specific application of the selected correction. \\
$\widetilde q(s)$ & Final calibrated membership score. \\
$\mathcal D^-_{\mathrm{cal}}$ & Known non-member calibration set. \\
$\mathcal E_{\mathrm{cand}}$ & Ten fixed candidate views. \\
$\mathcal E_{\mathrm{sel}}$ & Three views selected by FPP. \\
$u_{s,e}$ & Positive part of the detector-score shift. \\
$c_e$ & Per-view clipping cap. \\
$\mathbf H_e$ & Score-guided feature-shift matrix for view $e$. \\
$\mathbf U_{e,r}$ & Rank-$r$ shift basis estimated for view $e$. \\
$\mathbf G_r$ & Average projector for candidate rank $r$. \\
$\gamma_{r,i}$ & Cross-view consensus score of direction $\mathbf v_{r,i}$. \\
$\tau$ & Fixed cross-view consensus threshold. \\
$\mathbf B_r$ & Consensus basis containing directions with $\gamma_{r,i}\geq\tau$. \\
$\mathbf A_{r,\lambda}$ & Bounded correction matrix for rank $r$ and strength $\lambda$. \\
$\mathbf A^*$ & Correction matrix selected on calibration non-members. \\
$r$ & Per-view SVD rank. \\
$\lambda$ & Correction strength, $\lambda\in[0,1]$. \\
\bottomrule
\end{tabularx}
\end{table}

\section{End-to-end pipeline walkthrough}
\label{app:pipeline_walkthrough}

Algorithm~\ref{alg:calibdcd} gives the operational sequence of the two calibration stages.
Controlled views are used only to estimate and select $\mathbf A^*$. After selection, the same correction is applied to both supervised and reporting features, while the original query, feature extractor, classifier family, and training procedure remain unchanged.

\subsection{Numerical toy example}
\label{app:toy_example}

Suppose three known non-members have original scores $0.30$, $0.40$, and $0.60$ under threshold $0.50$.
Under a candidate view $v_1$, their scores become $0.45$, $0.70$, and $0.75$.
The positive score increases are $0.15$, $0.30$, and $0.15$, so
\begin{equation}
\fpp(v_1)=\frac{0.15+0.30+0.15}{3}=0.20.
\end{equation}

Suppose the corresponding raw feature shifts are
\[
[0.3,0.2,0],\quad [0.5,0.2,0],\quad [0.3,0.4,0].
\]
The $95$th-percentile cap of the three positive increases is $c_{v_1}=0.285$.
The effective weights are therefore $0.15$, $0.285$, and $0.15$.
After square-root weighting, the rows entering $\mathbf H_{v_1}$ are approximately
\[
\begin{aligned}
&[0.116,0.077,0],\quad [0.267,0.107,0],\\
&[0.116,0.155,0].
\end{aligned}
\]
The largest score increase is slightly clipped, so it remains influential without dominating the view-specific SVD.

Repeating the calculation for the other selected views gives their bases $\mathbf U_{e,r}$.
Assume the largest eigenvalues of the resulting consensus operator are $0.97$, $0.62$, and $0.31$.
With the fixed consensus threshold used in our experiments, the first direction is retained.
The second direction may be strong in part of the view set, but it does not receive sufficient cross-view support to define the correction.

\begingroup
\newcommand{\AlgStage}[1]{%
  \State {\color{green!45!black}\bfseries\itshape$\triangleright$~#1}%
}
\newcommand{\AlgNote}[1]{%
  \State {\color{green!45!black}\itshape$\triangleright$~#1}%
}
\begin{algorithm}[!t]
\footnotesize
\caption{\method two-stage calibration}
\label{alg:calibdcd}
\begin{algorithmic}[1]
\Statex \textbf{Input:} \parbox[t]{0.82\linewidth}{Target LLM $\Theta$, original query $e_0$, feature extractor $\phi$, detector score $q$, known non-member set $\mathcal D^-_{\mathrm{cal}}$, candidate views $\mathcal E_{\mathrm{cand}}$, rank grid $\mathcal R$, strength grid $\Lambda$, consensus threshold $\tau$}
\Statex \textbf{Output:} Selected correction matrix $\mathbf A^*$

\AlgStage{Stage 1: Multi-View Shift Detection.}
\AlgNote{Original-format reference.}
\State \textbf{Run} the detector under $e_0$ to obtain $\mathbf z_{s,e_0}$ and $q_{s,e_0}$

\AlgNote{Controlled-view shift estimation.}
\For{$e\in\mathcal E_{\mathrm{cand}}$}
  \State \textbf{Query} $\Theta$ under $e$ and extract $\mathbf z_{s,e}$ and $q_{s,e}$
  \State \textbf{Compute} $\Delta\mathbf z_{s,e}$, $\Delta q_{s,e}$, and $\fpp(e)$
\EndFor

\AlgNote{View selection.}
\State $\mathcal E_{\mathrm{sel}}\gets\operatorname{TopK}(\mathcal E_{\mathrm{cand}},\fpp,3)$

\AlgNote{Score-guided feature-shift construction.}
\For{$e\in\mathcal E_{\mathrm{sel}}$}
  \State $u_{s,e}\gets[\Delta q_{s,e}]_+$ and $c_e\gets\operatorname{Percentile}_{95}(\{u_{s,e}:u_{s,e}>0\})$
  \State \textbf{Form} $\mathbf H_e$ with rows $\sqrt{\min(u_{s,e},c_e)}\,\Delta\mathbf z_{s,e}$
\EndFor

\AlgNote{Cross-view consensus estimation.}
\For{$r\in\mathcal R$}
  \For{$e\in\mathcal E_{\mathrm{sel}}$}
    \State $\mathbf U_{e,r}\gets\operatorname{TopRightSV}(\mathbf H_e,r)$
  \EndFor
  \State $\mathbf G_r\gets |\mathcal E_{\mathrm{sel}}|^{-1}\sum_{e\in\mathcal E_{\mathrm{sel}}}\mathbf U_{e,r}\mathbf U_{e,r}^{\top}$
  \State $(\gamma_{r,i},\mathbf v_{r,i})_i\gets\operatorname{Eigh}(\mathbf G_r)$
  \AlgNote{Retain shared directions.}
  \State $\mathbf B_r\gets[\mathbf v_{r,i}:\gamma_{r,i}\geq\tau]$
\EndFor

\AlgStage{Stage 2: Bounded feature correction.}
\AlgNote{Correction-candidate evaluation.}
\For{$r\in\mathcal R$}
  \For{$\lambda\in\Lambda$}
    \State $\mathbf A_{r,\lambda}\gets\mathbf I-\lambda\mathbf B_r\mathbf B_r^{\top}$
    \State \textbf{Evaluate} $J(r,\lambda)$ using Equation~\ref{eq:parameter_selection}
  \EndFor
\EndFor

\AlgNote{Calibration-only selection.}
\State $(r^*,\lambda^*)\gets\arg\max_{r\in\mathcal R,\lambda\in\Lambda}J(r,\lambda)$
\State $\mathbf A^*\gets\mathbf A_{r^*,\lambda^*}$
\State \Return $\mathbf A^*$
\end{algorithmic}
\end{algorithm}
\endgroup

After calibration, the detector-specific transformation $\mathcal T_{\mathbf A^*}$ applies the selected correction to original-format supervised features before training $q_{\mathrm{final}}$ and to reporting features before evaluation.
The controlled views do not enter the post-calibration detection path.

\subsection{Workflow invariants}
\label{app:invariants}

Table~\ref{tab:invariants} summarizes the components held fixed throughout calibration and post-calibration detection.
The target LLM, original query, feature extractor, classifier family, supervised split, and training procedure remain unchanged.
Only the feature representation passed to the classifier is corrected.

\begin{table}[!htbp]
\centering
\small
\caption{Controlled comparison used by \method. The correction changes the feature vector passed to the classifier while preserving the detector query and classifier family.}
\label{tab:invariants}
\begin{tabularx}{\linewidth}{lY}
\toprule
\textbf{Component} & \textbf{Status under \method} \\
\midrule
Target LLM $\Theta$ & Unchanged. \\
Detection-time query format $e_0$ & Unchanged. \\
Feature extractor $\phi$ & Unchanged. \\
Classifier family / training procedure & Unchanged. \\
Original detector supervised split & Unchanged. Calibration texts are not added. \\
Ten controlled views & Used during calibration. \\
Feature vector $\mathbf z$ & Replaced by $\mathbf z'=\mathcal T_{\mathbf A^{*}}(\mathbf z)$. \\
\bottomrule
\end{tabularx}
\end{table}

\section{Fixed controlled views}
\label{app:view_bank}

\method uses ten fixed views for every dataset--model--detector setting.
The first eight views cover common response roles and output structures that recur across post-trained models.
The remaining two capture the assistant-generation boundary of the target model family and a nearby boundary variant.
The fixed $8+2$ design spans model-agnostic response conditions and model-specific assistant-generation boundaries.
It provides broad coverage of query conditions that can reveal post-training-induced feature changes.
The same ten view slots are reused across datasets, and FPP selects the three most informative views for each setting.

\begin{table}[!htbp]
\centering
\small
\caption{Eight universal controlled views. Each prefix is prepended to the same input text during calibration.}
\label{tab:universal_views}
\begin{tabularx}{\linewidth}{lY}
\toprule
\textbf{View} & \textbf{Prefix / behavior} \\
\midrule
Assistant role & \texttt{Assistant:} followed by a blank line. \\
User role & \texttt{User:} followed by a newline. \\
Response & \texttt{Response:} followed by a blank line. \\
Answer & \texttt{Answer:} followed by a blank line. \\
Reasoning & \texttt{Reasoning:} followed by a blank line. \\
Final answer & \texttt{Final answer:} followed by a blank line. \\
Summary & \texttt{Summary:} followed by a blank line. \\
Continuation & \texttt{Continue the text:} followed by a newline. \\
\bottomrule
\end{tabularx}
\end{table}

\begin{table}[!htbp]
\centering
\small
\caption{Two model-specific controlled views used with each target-model family.}
\label{tab:model_specific_views}
\begin{tabularx}{\linewidth}{lY}
\toprule
\textbf{Model family} & \textbf{Two assistant-boundary views} \\
\midrule
Qwen2.5 & Official assistant-generation boundary and the same boundary with an additional blank-line separator. \\
Llama 3.1 & Compact assistant header and the official assistant header with its standard blank-line separator. \\
DeepSeek-R1-Distill-Qwen & Assistant-role boundary without a reasoning opener and the official boundary followed by the reasoning opener. \\
\bottomrule
\end{tabularx}
\end{table}

\section{Consensus-score range}
\label{app:method_properties}

For a unit vector $\mathbf v$ and a matrix $\mathbf U_{e,r}$ with orthonormal columns,
$\|\mathbf U_{e,r}^{\top}\mathbf v\|_2^2$ is the squared norm of the projection of $\mathbf v$ onto the view-specific subspace and therefore lies in $[0,1]$.
Averaging across selected views gives
\[
0\leq
\frac{1}{|\mathcal E_{\mathrm{sel}}|}
\sum_{e\in\mathcal E_{\mathrm{sel}}}
\|\mathbf U_{e,r}^{\top}\mathbf v\|_2^2
\leq1.
\]
For an eigenvector $\mathbf v_{r,i}$ of $\mathbf G_r$, this average equals $\gamma_{r,i}$ by Equation~\ref{eq:consensus_score}. Therefore, $\gamma_{r,i}\in[0,1]$.

\section{Full configuration and hyperparameters}
\label{app:hyperparameters}

Table~\ref{tab:hyperparameters} lists the remaining configuration used in the experiments.
The cross-view consensus threshold and candidate grids are fixed in advance.
The strength grid includes partial correction with $0<\lambda<1$ and full correction with $\lambda=1$, which removes the retained consensus component.
The final rank--$\lambda$ pair is selected from the available calibration non-members by maximizing Equation~\ref{eq:parameter_selection}.
AUC, TPR@5\%FPR, and reporting-pool member labels do not participate in candidate selection.

\begin{table}[!htbp]
\centering
\scriptsize
\setlength{\tabcolsep}{2.5pt}
\renewcommand{\arraystretch}{0.96}
\caption{Full configuration. The lower block lists \method-specific parameters. The other rows are inherited from the underlying detector.}
\label{tab:hyperparameters}
\begin{tabularx}{\linewidth}{lY}
\toprule
\textbf{Parameter} & \textbf{Value} \\
\midrule
\multicolumn{2}{l}{\emph{Detector (inherited)}} \\
Original detector supervised split & $50$ members $+$ $50$ non-members \\
Episodes & $400$ \\
$n_{\mathrm{support}}$ & $10$ \\
$n_{\mathrm{query}}$ & $10$ \\
Normalization & z-score (supervised-split scope) \\
IB enabled & yes \\
IB $\beta$ & $0.005$ \\
Latent dimension & $128$ \\
Distance & squared Euclidean \\
Scoring & $\Delta=d(\tilde{\mathbf z},c_0)-d(\tilde{\mathbf z},c_1)$ \\
\midrule
\multicolumn{2}{l}{\emph{\method}} \\
Fixed candidate views & $10$ ($8$ universal $+$ $2$ model-specific) \\
FPP-selected views & $3$ \\
Known non-member calibration set & approximately half of the remaining known non-members outside the supervised split \\
Clipping cap $c_e$ & per-view $95$th percentile of positive $u_{s,e}$ values \\
Per-view SVD rank grid $\mathcal R$ & $\{3,4,5,6\}$ \\
Cross-view consensus rule & retain $\gamma_{r,i}\geq\tau$ \\
Cross-view consensus threshold $\tau$ & $0.95$ \\
Correction-strength grid $\Lambda$ & $0.7, 0.8, 0.9, 1.0$ \\
Number of rank--$\lambda$ candidates & $16$ \\
Candidate-selection signal & largest mean detector-score reduction on the available calibration non-members \\
Evaluation metrics used for selection & none \\
\bottomrule
\end{tabularx}
\end{table}

\begin{figure*}[!t]
\centering
\includegraphics[width=0.94\textwidth]{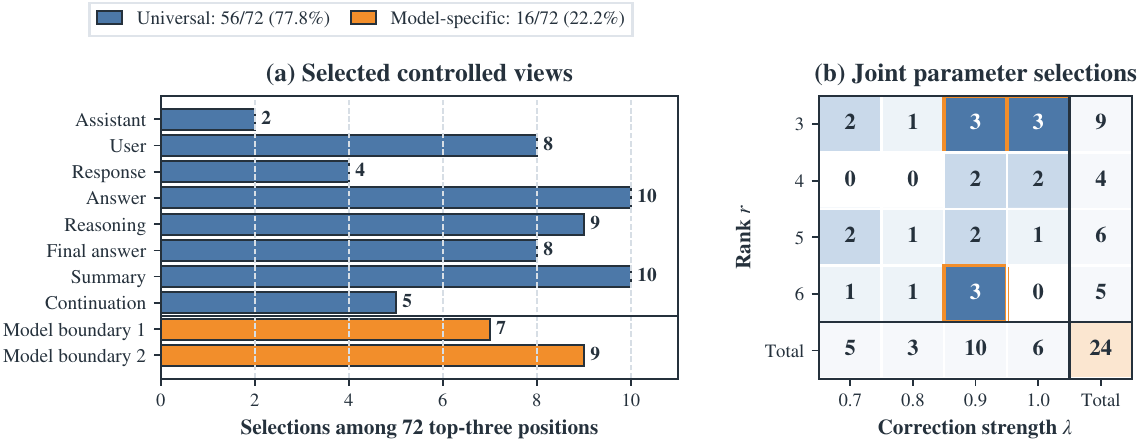}
\caption{Aggregate calibration choices across the $24$ benchmark--model--detector settings. (a) Selection frequencies across the $72$ top-three view positions. Category totals are reported in the legend, and the two model-specific slots aggregate the corresponding family-specific views. (b) Joint frequencies of the selected rank and correction strength, with row and column marginals. Orange outlines indicate the most frequent joint selections.}
\label{fig:selection_behavior}
\end{figure*}

\section{Base-vs-instruct configuration}
\label{app:base_vs_instruct}

Figure~\ref{fig:motivation} compares Qwen2.5-7B and Qwen2.5-7B-Instruct using VeilProbe on \textsc{BookTection}.
Table~\ref{tab:base_vs_instruct_config} reports the evaluated checkpoints and experimental scope.
The detector and data settings follow the inherited-detector block of Table~\ref{tab:hyperparameters}, making the target checkpoint the only difference.

\begin{table}[!htbp]
\centering
\scriptsize
\setlength{\tabcolsep}{2.5pt}
\renewcommand{\arraystretch}{0.96}
\caption{Base-vs-instruct comparison in Figure~\ref{fig:motivation}. Shared detector and data settings follow the inherited-detector block of Table~\ref{tab:hyperparameters}.}
\label{tab:base_vs_instruct_config}
\begin{tabularx}{\linewidth}{lY}
\toprule
\textbf{Item} & \textbf{Value} \\
\midrule
Base checkpoint & Qwen2.5-7B \\
Instruct checkpoint & Qwen2.5-7B-Instruct \\
Detector / benchmark & VeilProbe / \textsc{BookTection} \\
Reporting-pool $n$ & $1900$ \\
Shared detector and data settings & Identical to the inherited-detector block of Table~\ref{tab:hyperparameters}. \\
\bottomrule
\end{tabularx}
\end{table}

\section{Selection distributions}
\label{app:selection_behavior_analysis}

Figure~\ref{fig:selection_behavior}(a) reports the complete view-selection distribution across the $72$ top-three positions.
Universal views account for $56$ positions, while model-specific assistant-boundary views account for the remaining $16$.
No individual view is selected more than ten times, and at least one model-specific view is selected in $11$ settings.

Figure~\ref{fig:selection_behavior}(b) reports the joint distribution of the selected per-view rank and correction strength.
Every candidate rank and correction strength is selected in at least three settings.
Partial correction is selected in $18$ settings, and $\lambda=0.9$ is the most frequent strength.
No rank--strength pair appears more than three times.
These distributions show that the calibration choices vary across target models, benchmarks, and detector interfaces rather than collapsing to a single configuration.
Appendix~\ref{app:selected_configs} reports the complete setting-specific selections.

\section{Selected calibration configurations}
\label{app:selected_configs}

Tables~\ref{tab:selected_configs_vp} and~\ref{tab:selected_configs_dpdllm} report the rank, correction strength, and three FPP-selected views used for every setting.
Codes U1--U8 follow the universal-view order in Table~\ref{tab:universal_views}. Q1/Q2, L1/L2, and D2 denote the corresponding model-specific boundaries in Table~\ref{tab:model_specific_views}.
The configuration is selected separately for every benchmark--model--detector setting.

\begin{table}[H]
\centering
\footnotesize
\setlength{\tabcolsep}{2.5pt}
\renewcommand{\arraystretch}{0.98}
\caption{Selected VeilProbe configurations. The third column reports $r/\lambda$.}
\label{tab:selected_configs_vp}
\begin{tabularx}{\linewidth}{llcY}
\toprule
\textbf{Benchmark} & \textbf{Target} & \textbf{$r/\lambda$} & \textbf{Selected views} \\
\midrule
\textsc{BookTection} & Qwen & $4/1.0$ & U5, U8, U3 \\
 & Llama & $3/0.7$ & U5, U7, U2 \\
 & DeepSeek & $6/0.9$ & U8, U4, U7 \\
\textsc{BookMIA} & Qwen & $3/0.9$ & U6, Q2, Q1 \\
 & Llama & $4/0.9$ & U5, U3, U4 \\
 & DeepSeek & $3/0.7$ & U6, U4, U2 \\
\textsc{ArxivTection} & Qwen & $5/0.8$ & U2, U7, U4 \\
 & Llama & $4/0.9$ & U8, U5, U2 \\
 & DeepSeek & $6/0.8$ & U7, U2, U1 \\
\textsc{WikiMIA} & Qwen & $3/1.0$ & U5, U7, Q1 \\
 & Llama & $4/1.0$ & L2, L1, U2 \\
 & DeepSeek & $5/0.7$ & D2, U6, U4 \\
\bottomrule
\end{tabularx}
\end{table}

\begin{table}[H]
\centering
\footnotesize
\setlength{\tabcolsep}{2.5pt}
\renewcommand{\arraystretch}{0.98}
\caption{Selected DPDLLM configurations. The third column reports $r/\lambda$.}
\label{tab:selected_configs_dpdllm}
\begin{tabularx}{\linewidth}{llcY}
\toprule
\textbf{Benchmark} & \textbf{Target} & \textbf{$r/\lambda$} & \textbf{Selected views} \\
\midrule
\textsc{BookTection} & Qwen & $6/0.9$ & U5, U7, U6 \\
 & Llama & $3/0.9$ & L2, L1, U4 \\
 & DeepSeek & $3/0.9$ & D2, U7, U6 \\
\textsc{BookMIA} & Qwen & $3/1.0$ & U5, U7, U3 \\
 & Llama & $5/0.7$ & L2, L1, U6 \\
 & DeepSeek & $5/1.0$ & U3, U4, U6 \\
\textsc{ArxivTection} & Qwen & $3/0.8$ & U8, Q1, U2 \\
 & Llama & $3/1.0$ & L1, U5, L2 \\
 & DeepSeek & $5/0.9$ & U8, D2, U7 \\
\textsc{WikiMIA} & Qwen & $6/0.7$ & U6, U4, Q2 \\
 & Llama & $5/0.9$ & U2, U5, U4 \\
 & DeepSeek & $6/0.9$ & U7, U1, U4 \\
\bottomrule
\end{tabularx}
\end{table}